\documentclass[conference]{IEEEtran}
\IEEEoverridecommandlockouts
\usepackage{array}
\usepackage{booktabs} 
\usepackage{hyperref}
\hypersetup{
    colorlinks=false,   
    pdfborder={0 0 0}   
}

\usepackage{fontspec}
\usepackage{polyglossia}

\setdefaultlanguage{english}
\setotherlanguage{bengali}

\newfontfamily\bengalifont[
  Script=Bengali,
  Path= ./ ,
  Extension=.ttf
]{Kalpurush}
\usepackage{cite}
\usepackage{amssymb,amsfonts}
\usepackage{graphicx}
\usepackage{textcomp}
\usepackage{xcolor}
\usepackage{tikz}
\usepackage{nth}
\usepackage{subcaption} 
\usepackage{fancyhdr}
\usepackage{float}

\usepackage{algorithmicx,algpseudocode}
\usepackage[linesnumbered, ruled, vlined]{algorithm2e}

\usepackage{tabularx}
\usepackage{multirow}
\usepackage{threeparttable}
\usepackage{multicol}
\usepackage{xcolor,colortbl}
\usepackage{amsmath}
\usepackage{siunitx}

\usepackage{verbatim}

\usepackage{url}
\usepackage{textcomp}
\usepackage{upgreek}

\makeatletter
 \let\old@ps@headings\ps@headings
 \let\old@ps@IEEEtitlepagestyle\ps@IEEEtitlepagestyle
 \def\confheader#1{%

 \def\ps@IEEEtitlepagestyle{%
 \old@ps@IEEEtitlepagestyle%
 \def\@oddhead{\strut#1\hfill\strut}%
 \def\@evenhead{\strut\hfill#1\hfill\strut}%
 }%
 \ps@headings%
 }
 \makeatother

 \confheader{
    \noindent  
    \fontsize{9}{11}\selectfont  
    \parbox{\textwidth}{ 
        2025 International Conference on Electrical, Computer and Communication Engineering \\ 
        (ECCE), 13-15 February 2025, CUET, Chattogram-4349, Bangladesh
    }
}

  \usepackage[pscoord]{eso-pic}
 \newcommand{\placetextbox}[3]{
  \setbox0=\hbox{#3}
 \AddToShipoutPictureFG*{ \put(\LenToUnit{#1\paperwidth},\LenToUnit{#2\paperheight}){\vtop{{\null}\makebox[0pt][c]{#3}}}
  }
  }
  \placetextbox{.50}{0.08}{\small{Authors' version of the paper published in proceedings of ECCE 2025, DOI: https://doi.org/10.1109/ECCE64574.2025.11012950}}
    
\begin{document}
\title{Bridging the Gap in Bangla Healthcare: Machine Learning Based Disease Prediction Using a Symptoms-Disease Dataset}

 \author {\IEEEauthorblockN{Rowzatul Zannat\IEEEauthorrefmark{1}, Abdullah Al Shafi\IEEEauthorrefmark{2}, and Abdul Muntakim\IEEEauthorrefmark{3}}
 \IEEEauthorblockA{Department of Computer Science and Engineering\\
 Khulna University of Engineering \& Technology, Khulna-9203, Bangladesh\\
 w.rzrowza@gmail.com\IEEEauthorrefmark{1}, abdullahratulk@gmail.com\IEEEauthorrefmark{2}, basitmuntakim@gmail.com\IEEEauthorrefmark{3}}}

\maketitle
\begin{abstract}
Increased access to reliable health information is essential for non-English-speaking populations, yet resources in Bangla for disease prediction remain limited. This study addresses this gap by developing a comprehensive Bangla symptoms-disease dataset containing 758 unique symptom-disease relationships spanning 85 diseases.  To ensure transparency and reproducibility, we also make our dataset publicly available. The dataset enables the prediction of diseases based on Bangla symptom inputs, supporting healthcare accessibility for Bengali-speaking populations. Using this dataset, we evaluated multiple machine learning models to predict diseases based on symptoms provided in Bangla and analyzed their performance on our dataset. Both soft and hard voting ensemble approaches combining top-performing models achieved 98\% accuracy, demonstrating superior robustness and generalization. Our work establishes a foundational resource for disease prediction in Bangla, paving the way for future advancements in localized health informatics and diagnostic tools. This contribution aims to enhance equitable access to health information for Bangla-speaking communities, particularly for early disease detection and healthcare interventions.
\end{abstract}


\section{Introduction}
Information searches are becoming more time-consuming and expensive due to the need for health-related data. There are billions of searches per day, and thousands of results about health advice are generated\cite{article}\cite{HONG2018175}. By discovering important trends and connections between various variables in vast databases, using a range of machine learning techniques\cite{article3} and data mining tools\cite{article2} has revolutionized healthcare organizations.

However, a notable gap exists for non-English speakers. Bangla, one of the most spoken languages in the world\footnote{https://en.wikipedia.org/wiki/Bengali\_language}, has limited representation in healthcare data applications\cite{article4}. This language gap restricts millions of Bangla speakers from accessing reliable health insights, highlighting the need for inclusive tools that support diverse linguistic groups. Bridging this gap would open up essential healthcare resources to Bangla-speaking communities, making medical information more accessible and ensuring more equitable healthcare for millions.

Machine learning algorithms\cite{article3} and deep learning techniques\cite{deep} have significantly advanced disease prediction, with models such as decision trees \cite{hashi2017expert}\cite{faruque2019performance}, support vector machines (SVM)\cite{arumugam2023multiple}, random forests \cite{grampurohit2020disease}, and neural networks\cite{gallant1990perceptron} achieving high accuracy in diagnosing conditions like hepatitis, diabetes, dengue, HIV/AIDS, pneumonia, rheumatism. These systems identify patterns and early signs of diseases, allowing for timely and personalized healthcare interventions.

Data mining combined with Machine Learning\cite{dataml} allows early disease identification and increased diagnostic precision, essential to the healthcare industry\cite{article2} \cite{article3}. By identifying potential diseases at an early stage, we can facilitate timely treatment and also make it easier for healthcare professionals to provide the necessary care to patients. Hepatitis, diabetes, dengue, HIV/AIDS, pneumonia, rheumatism, and other illnesses have a serious negative influence on health and can be lethal if neglected. Although they lack easily accessible resources to offer precise insights, people frequently wish to know if their symptoms point to a serious illness. The medical industry may benefit greatly from machine learning technology\cite{article3}, which makes it possible to treat healthcare issues more effectively and gives people vital information for prompt response.

In our experiment, we aimed to accurately predict diseases by analyzing patients' symptoms in Bangla. To address the scarcity of notable work for disease prediction and the lack of publicly available symptoms-disease corpora in Bangla, we created a Bangla symptoms-disease dataset\footnote{https://data.mendeley.com/datasets/rjgjh8hgrt/5} and evaluated the effectiveness of several machine learning models before combining the top three models to create soft and hard voting ensembles that would allow for greater generalization.

The introduction to general disease prediction using classification techniques is covered in section I. The literature evaluation of current systems is presented in Section II, and our developed dataset of symptoms-disease, along with the specifics of the disease, symptoms, and their relationships, is presented in Section III. Section IV presents our proposed disease prediction approach. Section V contains the analysis and findings of the experiment. In section VI, a conclusion is finally drawn.

\section{Literature Review}
In recent years, studies on disease prediction have utilized a variety of datasets and machine learning models to enhance prediction accuracy. Sneha Grampurohit et al.\cite{grampurohit2020disease} and Md. Atikur et al.\cite{rahman2023predicting} utilized a popular dataset from Kaggle\footnote{https://www.kaggle.com/datasets/kaushil268/disease-prediction-using-machine-learning/data}, which contains 4,920 records and can predict 41 distinct diseases. This dataset has also been adapted in Bangla-language research by M. M. Rahman et al.\cite{rahman2019disha}, who translated the data into Bangla using the Google Translation API, thus making it accessible for localized disease prediction studies.

Dhiraj Dahiwade et al.\cite{dahiwade2019designing} applied datasets from the UCI Machine Learning Repository specifically to explore diabetes prediction. In a more region-specific approach, Faisal Faruque et al.\cite{faruque2019performance} gathered information on diabetes from Bangladesh's Diagnostic Medical Centre of Chittagong (MCC), offering valuable insights into diabetes prediction within the Bangladeshi population.

Machine learning models such as decision tree, random forest, naive bayes, and SVM have been widely tested, consistently achieving high accuracy across various datasets. For instance, Sneha Grampurohit and Chetan Sagarnal\cite{grampurohit2020disease} trained models using a decision tree, random forest, and naive bayes, achieving a test accuracy of 95.12\%. Similarly, K. Arumugam et al.\cite{arumugam2023multiple} compared decision tree, naive bayes and SVM, finding that decision tree achieved the highest accuracy.

In a Bangla-language study on disease prediction, researchers\cite{rahman2019disha} tested a range of models, including decision tree, KNN, random forest, naive bayes, SVM and AdaBoost, with SVM reaching the highest accuracy at 98.39\%, underscoring its effectiveness for Bangla datasets.

In order to overcome the lack of resources in Bangla for disease prediction, we created a Bangla symptoms-disease dataset in this research. This dataset was used to train and test eight different machine learning algorithms for accurate predictions and to evaluate their performance. After that we design soft and hard voting ensembles of the best three models, proving especially effective for our dataset. 

\section{A Bangla Symptoms-Disease Dataset}
To effectively classify diseases based on their symptoms, our first crucial step was to develop a robust dataset. The process began with extensive research and data collection, which included conducting comprehensive surveys within the medical community to gather authentic and reliable insights directly from professionals. Additionally, we explored various online data sources, such as medical journals, health-related blogs, newspaper articles, and forums where symptom-related experiences were shared. By integrating the findings from both approaches, we meticulously compiled a raw data set that was cleaned and pre-processed to ensure quality and relevance. As a result, we now have a well-structured and comprehensive dataset ready to be used for training and testing our classification models. We have also made our dataset publicly available\footnote{https://data.mendeley.com/datasets/rjgjh8hgrt/5}.

\begin{table}[ht]
\caption{An overview of the source test and train datasets}
\centering
\begin{tabular}{|l|c|c|c|}
\hline
 \textbf{Dataset} &\textbf{Total}& \textbf{Train} & \textbf{Test} \\
\hline
\textbf{Sample Size} & 758 & 607 & 151 \\
\hline
\textbf{Unique Diseases} & 85 & 85 & 85\\
\hline
\end{tabular}
\label{tab:dataset_summary}
\end{table}

Table \ref{tab:dataset_summary} shows our developed dataset in brief. There are 85 unique diseases, 172 symptoms, and a total 758 relation between them i.e. sample size.

\begin{table*}[!ht]
\caption{Diseases in our dataset} 
\centering 
\small 
\begin{tabular}{|c|c|c|} 
\hline 
\multicolumn{3}{|c|}{\textbf{Diseases} }\\ 
\hline 
\textbengali{ফাঙ্গাস (Fungal infection)} & \textbengali{ডায়েবেটিস (Diabetes)} & \textbengali{ক্যান্সার (Cancer)} \\ 
\hline 
\textbengali{ডিমনেশিয়া (Dementia)} & \textbengali{স্ট্রোক (Stroke)} & \textbengali{ডেঙ্গু (Dengue)} \\ 
\hline 
\textbengali{আমাশয় (Dysentery)} & \textbengali{জলবসন্ত (Chickenpox)} & \textbengali{কোষ্ঠকাঠিন্য (Constipation)} \\ 
\hline 
\textbengali{চোখ ওঠা (Conjunctivitis)} & \textbengali{গুটিবসন্ত (Smallpox)} & \textbengali{দাদ রোগ (Ringworm)} \\ 
\hline 
\textbengali{বাত (Gout)} & \textbengali{কলেরা (Cholera)} & \textbengali{নিমোনিয়া (Pneumonia)} \\ 
\hline 
\textbengali{ম্যালেরিয়া (Malaria)} & \textbengali{হার্ট অ্যাটাক (Heart Attack)} & \textbengali{বধিরতা (Deafness)} \\ 
\hline 
\textbengali{পারকিনসন্স (Parkinson's Disease)} & \textbengali{গ্লুকোমা (Glaucoma)} & \textbengali{ছানি (Cataract)} \\ 
\hline 
\textbengali{বেরিবেরি (Beriberi)} & \textbengali{রিকেট (Rickets)} & \textbengali{জলাতঙ্ক (Rabies)} \\ 
\hline 
\textbengali{সোয়াইন ফ্লু (Swine Flu)} & \textbengali{পক্ষাঘাত (Paralysis)} & \textbengali{এপিলেপসি (Epilepsy)} \\ 
\hline 
\textbengali{নিপাহ ভাইরাস (Nipah Virus)} & \textbengali{ধনুষ্টঙ্কার (Tetanus)} & \textbengali{ওটিটিস মিডিয়া (Otitis Media)} \\ 
\hline 
\textbengali{গ্লসাইটিস (Glossitis)} & \textbengali{কুষ্ঠ (Leprosy)} & \textbengali{চিকুনগুনিয়া (Chikungunya)} \\ 
\hline 
\textbengali{আলসার (Ulcer)} & \textbengali{ইনফ্লুয়েঞ্জা (Influenza)} & \textbengali{হাম (Measles)} \\ 
\hline 
\textbengali{হিমোফিলিয়া (Hemophilia)} & \textbengali{স্কার্ভি (Scurvy)} & \textbengali{প্লুরিসি (Pleurisy)} \\ 
\hline 
\textbengali{পায়োরিয়া (Pyorrhea)} & \textbengali{ট্র্যাকোমা (Trachoma)} & \textbengali{সাইনোসাইটিস (Sinusitis)} \\ 
\hline 
\textbengali{ব্রংকাইটিস (Bronchitis)} & \textbengali{পোলিও (Polio)} & \textbengali{স্ক্লেরোসিস (Sclerosis)} \\ 
\hline 
\textbengali{মস্তিষ্কের টিউমার (Brain Tumor)} & \textbengali{এনসেফালাইটিস (Encephalitis)} & \textbengali{সিসিএইচএফভি (CHIKV)} \\ 
\hline 
\textbengali{লিউকোমিয়া (Leukemia)} & \textbengali{ডিপথেরিয়া (Diphtheria)} & \textbengali{যক্ষ্মা (Tuberculosis)} \\ 
\hline 
\textbengali{টাইফয়েড (Typhoid)} & \textbengali{জন্ডিস (Jaundice)} & \textbengali{গলগণ্ড (Goiter)} \\ 
\hline 
\textbengali{রিফট ভ্যালি ফিভার (Rift Valley Fever)} & \textbengali{এলার্জি (Allergy)} & \textbengali{জিইআরডি (GERD)} \\ 
\hline 
\textbengali{ক্রনিক কোলেস্টেসিস (Chronic Cholecystitis)} & \textbengali{ওষুধের প্রতিক্রিয়া (Drug Reaction)} & \textbengali{পেপটিক আলসার (Peptic Ulcer)} \\ 
\hline 
\textbengali{এইডস (AIDS)} & \textbengali{গ্যাস্ট্রোএন্টেরাইটিস (Gastroenteritis)} & \textbengali{ব্রঙ্কিয়াল অ্যাজমা (Bronchial Asthma)} \\ 
\hline 
\textbengali{উচ্চ রক্তচাপ (Hypertension)} & \textbengali{মাইগ্রেন (Migraine)} & \textbengali{সার্ভিকাল স্পন্ডাইলোসিস (Cervical Spondylosis)} \\ 
\hline 
\textbengali{প্যারালাইসিস (Paralysis)} & \textbengali{হেপাটাইটিস এ (Hepatitis A)} & \textbengali{হেপাটাইটিস বি (Hepatitis B)} \\ 
\hline 
\textbengali{হেপাটাইটিস সি (Hepatitis C)} & \textbengali{হেপাটাইটিস ডি (Hepatitis D)} & \textbengali{হেপাটাইটিস ই (Hepatitis E)} \\ 
\hline 
\textbengali{অ্যালকোহলিক হেপাটাইটিস (Alcoholic Hepatitis)} & \textbengali{সাধারণ সর্দি (Common Cold)} & \textbengali{পাইলস(Piles)} \\ 
\hline 
\textbengali{ভ্যারিকোস ভেইন (Varicose Veins)} & \textbengali{হাইপোথাইরয়েডিজম (Hypothyroidism)} & \textbengali{হাইপারথাইরয়েডিজম (Hyperthyroidism)} \\ 
\hline 
\textbengali{হাইপোগ্লাইসেমিয়া (Hypoglycemia)} & \textbengali{অস্টিওআর্থারাইটিস (Osteoarthritis)} & \textbengali{ভারটিগো (Vertigo)} \\ 
\hline 
\textbengali{ব্রণ (Acne)} & \textbengali{মূত্রনালীর সংক্রমণ (Urinary Tract Infection)} & \textbengali{সোরিয়াসিস (Psoriasis)} \\ 
\hline 
\textbengali{ইমপেটিগো (Impetigo)} & & \\ 
\hline 
\end{tabular} 
\label{tab:disease_table} 
\end{table*}

The list of diseases covered in our dataset is shown in table \ref{tab:disease_table} which contains 85 unique diseases, illustrating the range and diversity of diseases in our dataset.

\begin{figure}[htbp]
\centering
\centerline{\includegraphics[scale=.40]{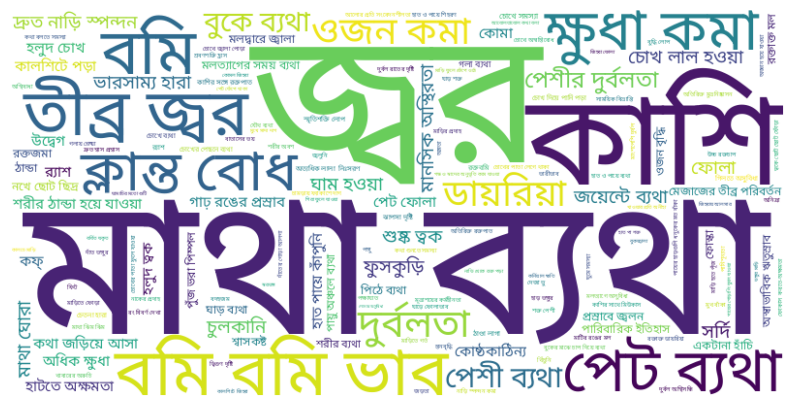}}
\caption{Word Cloud of Symptoms in our dataset}
\label{fig:symptoms_cloud}
\end{figure}

Fig. \ref{fig:symptoms_cloud} shows the word cloud of the symptoms that are covered in our dataset, visually illustrating the range and diversity of symptoms in our dataset.

\begin{table*}[!ht]
\caption{Sample Dataset}
\centering
\begin{tabular}{|c|c|c|c|c|c|c|c|c|c|c|c|c|c|} 
\hline
\multicolumn{13}{|c|}{Symptoms} & \multirow{2}{*}{Disease}\\
\cline{1-13}
 \bengalifont{শ্বাসকষ্ট} & \bengalifont{তীব্র জ্বর} & \bengalifont{পেট ব্যথা} &
 \bengalifont{...}  & 
 \bengalifont{বমি} &
 \bengalifont{দুর্বলতা} &
 \bengalifont{শরীর ব্যথা} &
  \bengalifont{ফুসকুড়ি} &
 \bengalifont{র‌্যাশ}  & \bengalifont{অজ্ঞান}  &
 \bengalifont{বুকে ব্যথা}  &
 \bengalifont{কাশি}  &
 \bengalifont{অজ্ঞান}&  
  \\
\hline
\bengalifont{1} & \bengalifont{1} & \bengalifont{0} &
\bengalifont{...}  & 
\bengalifont{0} &
\bengalifont{0}  &
\bengalifont{0} &
\bengalifont{0} &
\bengalifont{1} &
\bengalifont{0} &
\bengalifont{0} &
\bengalifont{0} &
\bengalifont{1} & \bengalifont{ডেঙ্গু (Dengue)}\\
  \hline
\bengalifont{0} & \bengalifont{0} & 
\bengalifont{1} &
\bengalifont{...}  & 
\bengalifont{1} &
\bengalifont{1} &
 
\bengalifont{0} &
\bengalifont{0} &
\bengalifont{0} &
\bengalifont{0} &
\bengalifont{0} &
\bengalifont{0} &
\bengalifont{0} & \bengalifont{আমাশয় (Dysentery)}\\
 \hline
\bengalifont{1} & \bengalifont{1} & 
\bengalifont{0} &
\bengalifont{...}  & 
\bengalifont{0} &
\bengalifont{0} &
\bengalifont{1} & 
\bengalifont{0} &
\bengalifont{0} &
\bengalifont{0}  & \bengalifont{1} &
\bengalifont{1} &
\bengalifont{0} &
\bengalifont{নিমোনিয়া (Pneumonia)}\\
 \hline
\bengalifont{0} & \bengalifont{1} & 
\bengalifont{0} &
\bengalifont{...}  & 
\bengalifont{0} &
\bengalifont{1} &
 \bengalifont{1} &
\bengalifont{0} &
\bengalifont{0} &
\bengalifont{0}  & \bengalifont{0} &
\bengalifont{0} &
\bengalifont{0} &\bengalifont{ম্যালেরিয়া (Malaria)}\\
 \hline
\bengalifont{1} & \bengalifont{1} & 
\bengalifont{0} &
\bengalifont{...}  & 
\bengalifont{0} &
\bengalifont{0} &
 \bengalifont{1} &
\bengalifont{0} &
\bengalifont{0} &
\bengalifont{0}  & \bengalifont{1} &
\bengalifont{1} &
\bengalifont{0} &
\bengalifont{ব্রংকাইটিস (Bronchitis)}\\
 \hline
\end{tabular}
\label{tab:sample_table}
\end{table*}

Table \ref{tab:sample_table} shows a sample portion from our symptoms-disease dataset. The disease prediction system uses a binary feature dataset with columns representing symptoms, with \texttt{'}1\texttt{'} indicating presence and \texttt{'}0\texttt{'} indicating absence. The final column, \texttt{'}Disease\texttt{'} contains the disease name based on the symptom pattern, enabling the prediction of diseases based on their presence or absence.

\section{Proposed Methodology}
\begin{figure}[htbp]
\centering
\centerline{\includegraphics[scale=.55]{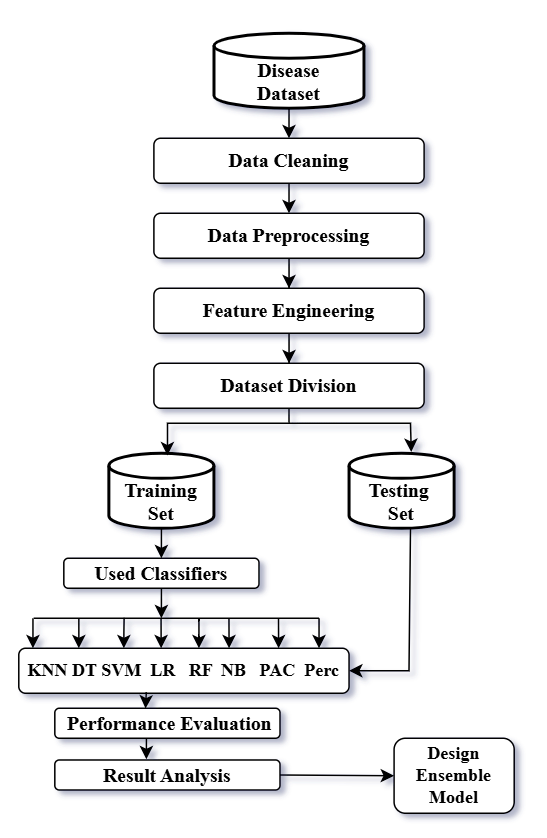}}
\caption{The proposed methodology in brief}
\label{fig:proposed}
\end{figure}

The input symptoms undergo a data cleaning process, following several pre-processing steps. After that, several classification algorithms are applied to categorize the combination of symptoms of a disease. Fig. \ref{fig:proposed} shows the proposed methodology in brief.

\subsection{Data Cleaning}
Identifying and correcting errors, gaps, irregularities, and odd structures in datasets to make them more illuminating and suitable for analysis and significant interpretation is known as data cleaning. In our dataset, we have 172 symptoms with 85 diseases. As a combination of a few symptoms is mapped to a disease of a row, there is a lot of 0 in the dataset. There may have been an overlooked 0 in some cells. To handle these missing values, we manipulate the dataset by putting 0 to those values.

\subsection{Preprocessing}
Feature encoding, target encoding, feature scaling, and dimensionality reduction are applied in our task as part of preprocessing.
\subsubsection{Feature Encoding}
One-hot encoding is applied for feature(symptom) encoding. For most of the models in our work, one-hot encoding is effective for categorical symptoms as it provides a non-ordinal structure.
\subsubsection{Target Encoding}
We apply label encoding for target(disease) encoding. As the target variable is a single categorical label, label encoding is straightforward. The target variable is treated as a discrete category without implying any ordinal relationship.
\subsubsection{Feature Scaling}
We standardize features to zero mean and unit variance. This is beneficial for many models in our work, as they perform better with scaled data.
\subsubsection{Dimensionality Reduction}
We use Principal Component Analysis(PCA) to reduce dimensions, keeping 90\% of the explained variance. As the dataset has 172 features (high dimensional data), PCA has proved to be beneficial to reduce noise and computational costs.

\subsection{Classification Algorithms}
\subsubsection{Perceptron}
Perceptron\cite{gallant1990perceptron} is a basic linear classifier that adjusts weights in response to prediction errors, works only with linearly separable data, and serves as the basis for neural networks. 
\subsubsection{Logistic Regression}
Simple and effective for linearly separable data, linear regression\cite{schober2021logistic} calculates the likelihood of a class based on feature weights and bias.
\subsubsection{Naive Bayes}
Naive bayes\cite{wickramasinghe2021naive} is a probabilistic model that assumes independence between features, making it efficient and effective for classification tasks.
\subsubsection{Decision Tree}
Using tree-like models of a decision based on feature splits, Decision tree\cite{hashi2017expert} is effective in capturing non-linear relationships.
\subsubsection{K-Nearest Neighbors}
KNN\cite{rahman2019disha} classifies based on the majority label among the nearest data points and performs well on small to moderate datasets.
\subsubsection{Passive Aggressive Classifier}
Being an online learning that adjusts only when predictions are incorrect, PAC\cite{crammer2006online} is suitable for large, streaming datasets.
\subsubsection{Random Forest}
Random forest\cite{grampurohit2020disease} is an ensemble of decision trees. It improves model performance by averaging results from multiple trees, thus reducing overfitting.
\subsubsection{Support Vector Machine}
Using kernels to handle non-linear data, SVM\cite{arumugam2023multiple} finds the best hyperplane to maximally separate classes.

\subsection{Baseline Model Validation}
The baseline performance of each model on the training set is assessed using 5-fold cross-validation. Prior to tuning, it is advantageous to set the hyperparameters' initial values.

\subsection{Hyperparameter tuning with cross validation}
In hyperparameter tuning, to determine the most suitable parameters, we employ 5-fold cross-validation. Table \ref{tab:models_parameters} displays the optimal hyperparameters for each model, where C is the parameter for regularization.

\begin{table}[h!]
\centering
\caption{Models with their best hyperparameters}
\begin{tabular}{|l|l|l|}
\hline
\textbf{Models} & \textbf{Hyperparameters} & \textbf{Best Value} \\ \hline
Perceptron & alpha & 0.001 \\ \cline{2-3} & penalty & l1 \\ \hline
Logistic Regression  & C & 0.1 \\ \hline
Decision Tree & min samples leaf & 2 \\ \cline{2-3} & max depth & 65 \\ \hline
K-Nearest Neighbors & n\_neighbors & 5 \\ \cline{2-3} & weights & uniform \\  \hline
PAC & C & 0.01 \\ \cline{2-3} & loss & squared hinge \\ \hline
Random Forest & n\_estimators & 70 \\ \cline{2-3} & max depth & 50 \\ \cline{2-3} & min samples leaf & 2 \\ \hline
Support Vector Machine & kernel & linear \\ \cline{2-3} & C & 0.1 \\ \hline
\end{tabular}
\label{tab:models_parameters}
\end{table}

\subsection{Ensemble Methods}
We combine the output of the top three performing methods to make soft and hard voting ensemble models.
\subsubsection{Soft Voting Ensemble}
The approach being used in the soft voting ensemble is probabilistic. The final prediction is calculated by averaging the probabilities for each class label in each classification algorithm.
\begin{equation}
P_j = \frac{1}{M} \sum_{i=1}^{M} P_{i,j}
\end{equation}

\begin{equation}
\hat{y} = \arg \max_{j \in \{1, 2, \dots, k\}} P_j
\end{equation}

\( P_{i,j} \) is the probability of class \( j \) predicted by classification algorithm \( C_i \) and \( k \) is the number of classes.

\subsubsection{Hard Voting Ensemble}
Every classification algorithm in hard voting ensemble predicts a distinct class and the class with the highest number of votes from all algorithms is the final class.
\begin{equation}
V_j = \sum_{i=1}^{M} \mathbf{1}_{\{\hat{y}_i = j\}}
\end{equation}

\begin{equation}
\hat{y} = \arg \max_{j \in \{1, 2, \dots, k\}} V_j
\end{equation}

\( \hat{y}_i \) is the predicted class by classification algorithm \( C_i \) for a given sample, \( \mathbf{1}_{\{\hat{y}_i = j\}} \) is an indicator function that equals 1 if classifier \( C_i \) predicts class \( j \), and 0 otherwise and \( k \) is the number of classes.

\begin{table*}[!ht]
\centering
\setlength\extrarowheight{6pt}
\label{result}
\caption{Result on test set of our disease prediction dataset} 
\begin{tabular}{|c|c|cc|cc|cc|}
\hline
\multirow{2}{*}{\textbf{Model}} &
  \multirow{2}{*}{\textbf{Accuracy}} &
  \multicolumn{2}{c|}{\textbf{Precision}} &
  \multicolumn{2}{c|}{\textbf{Recall}} &
  \multicolumn{2}{c|}{\textbf{F1-score}} \\ \cline{3-8} 
 &
   &
  \multicolumn{1}{c|}{\textbf{Macro}} &
  \textbf{Micro} &
  \multicolumn{1}{c|}{\textbf{Macro}} &
  \textbf{Micro} &
  \multicolumn{1}{c|}{\textbf{Macro}} &
  \textbf{Micro} \\ \hline
Perceptron  & \textbf{0.97} & \multicolumn{1}{c|}{0.93} & 0.97  & \multicolumn{1}{c|}{\textbf{0.92}} & 0.97 & \multicolumn{1}{c|}{\textbf{0.92}} & 0.97  \\ \hline
Logistic Regression & \textbf{0.97} & \multicolumn{1}{c|}{0.96} & 0.97 & \multicolumn{1}{c|}{\textbf{0.95}} & 0.97 & \multicolumn{1}{c|}{\textbf{0.95}}  & 0.97 \\ \hline
Naive Bayes & 0.94 & \multicolumn{1}{c|}{0.91} & 0.94 & \multicolumn{1}{c|}{0.90} & 0.94 & \multicolumn{1}{c|}{0.90}  & 0.94 \\ \hline
Decision Tree  & 0.78 & \multicolumn{1}{c|}{0.71} & 0.78 & \multicolumn{1}{c|}{0.70} & 0.78 & \multicolumn{1}{c|}{0.69} & 0.78 \\ \hline
K-Nearest Neighbors & 0.96 & \multicolumn{1}{c|}{0.93} & 0.96 & \multicolumn{1}{c|}{0.91} & 0.96 & \multicolumn{1}{c|}{0.91} & 0.96 \\ \hline
Passive Aggressive Classifier & 0.96 & \multicolumn{1}{c|}{0.91} & 0.96 & \multicolumn{1}{c|}{0.90} & 0.96 & \multicolumn{1}{c|}{0.90} & 0.96 \\ \hline
Random Forest   & \textbf{0.97} & \multicolumn{1}{c|}{0.96} & 0.97 & \multicolumn{1}{c|}{\textbf{0.96}} & 0.97 & \multicolumn{1}{c|}{\textbf{0.96}}   & 0.97 \\ \hline
Support Vector Machine  & 0.96  & \multicolumn{1}{c|}{0.94} & 0.96 & \multicolumn{1}{c|}{0.92} & 0.96 & \multicolumn{1}{c|}{0.92} & 0.96 \\ \hline
Soft Voting Ensemble  &  \textbf{0.98}  & \multicolumn{1}{c|}{0.97} & 0.98 & \multicolumn{1}{c|}{\textbf{0.97}} & 0.98 & \multicolumn{1}{c|}{\textbf{0.97}}  & 0.98 \\ \hline
Hard Voting Ensemble  &  \textbf{0.98}  & \multicolumn{1}{c|}{0.97} & 0.98 & \multicolumn{1}{c|}{\textbf{0.97}} & 0.98 & \multicolumn{1}{c|}{\textbf{0.97}}  & 0.98 \\ \hline
\end{tabular}
\label{tab:result}
\end{table*}

\section{Experimental Results and Analysis}

\subsection{Train-Test Split}
The percentage of data for training and testing is selected at random and varies based on task requirements and available data to ensure that data is adequately represented in both training and test sets. 80\% of the data was utilized for training, while 20\% was used for testing.

\subsection{Evaluation Metrics}
Accuracy, precision, recall, and F1-score are crucial parameters to take into account while assessing disease prediction models. When dealing with a skewed dataset, it is imperative to take bias into account. In disease prediction, recall is generally prioritized as the goal is to minimize the chance of missing any disease entirely. Since the macro-recall score provides a more fair evaluation of the model's effectiveness in every class and is able to detect biases that the micro-recall score might not be able to easily identify, it is a more suitable metric for disease prediction model evaluation on biased datasets. Macro-F1 score is also important to make a balance between not missing a disease and misdiagnosing someone with the wrong disease.

\subsection{Results}


Table \ref{tab:result} lists the various performance metrics for every model used in the research. With a macro-recall of 97\%, macro-F1 score of 97\%, and accuracy of 98\%, ensemble methods (soft and hard voting) appear to be the best performers among all the models. By utilizing many classifiers, these models were more robust to the different types of errors that individual models may produce. Because of their superior performance, they ought to be given priority for this task.\\
Random forest showed better performance with a macro-recall of 96\%, macro-F1 score of 96\%, and accuracy of 97\%. Because of being robust to overfitting, the model generalizes well to the disease prediction task. Logistic regression and perceptron also performed quite well. KNN and SVM were also competitive. Naive Bayes had lower performance compared to the previous models, possibly due to the violation of its independence assumption. The decision tree was by far the worst performer.

\begin{figure}[htbp]
\centering
\centerline{\includegraphics[scale=.60]{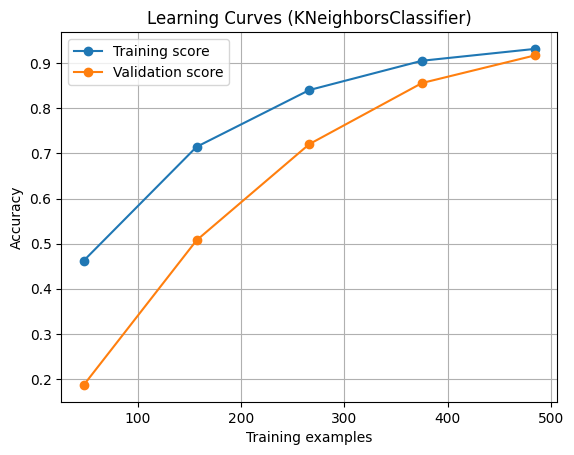}}
\caption{Learning curve of K-Nearest Neighbors. The convergence pattern demonstrates a healthy learning behavior for KNN and shows a well-fitted model that is neither underfitting nor overfitting.}
\label{knn_learning_curve}
\end{figure}

Fig. \ref{knn_learning_curve} shows the learning curve of KNN. The model's accuracy on the training set begins relatively low with only a few examples and gradually improves, reaching about 0.95 when all examples are used. This suggests that the model is better able to learn and represent the underlying data distribution as more data becomes available. Model accuracy on unseen data is represented by the validation accuracy. The validation accuracy continuously increases with the size of the training set, suggesting that the model is picking up more relevant patterns.\\
The model gets better at generalizing with more data, and the gap between training and validation accuracy gets less, as evidenced by the convergence of training and validation scores as the dataset size grows. The same goes for other models used in our study.

\begin{table}[h!]
\centering
\caption{Qualitative Analysis of Our Study. Here, the inputs symptoms are \textbengali{হলুদ ত্বক, গাঢ় রঙের প্রস্রাব, চুলকানি}(Yellowish skin, dark urine, itching) and the actual disease is  \textbengali{জন্ডিস}(Jaundice)}.
\label{tab:qualitative}
\begin{tabular}{|c|c|}
\hline
\textbf{Method} & \textbf{Prediction}\\
\hline
Perceptron & \textbengali{জন্ডিস (Jaundice)} \\
\hline
Logistic Regression & \textbengali{জন্ডিস (Jaundice)} \\
\hline
Naive Bayes & \textbengali{জন্ডিস (Jaundice)} \\
\hline
Decision Tree & \textbengali{\color{red}স্ক্লেরোসিস (Sclerosis)} \\
\hline
KNN & \textbengali{জন্ডিস (Jaundice)} \\
\hline
PAC & \textbengali{জন্ডিস (Jaundice)} \\
\hline
Random Forest & \textbengali{জন্ডিস (Jaundice)} \\
\hline
SVM & \textbengali{জন্ডিস (Jaundice)} \\
\hline
Soft Voting Ensemble & \textbengali{জন্ডিস (Jaundice)} \\
\hline
Hard Voting Ensemble & \textbengali{জন্ডিস (Jaundice)} \\
\hline
\end{tabular}
\end{table}

Table \ref{tab:qualitative} shows the qualitative analysis of various methods we have used. It shows that all the models can correctly predict the disease except the decision tree.






\section{Conclusions}
In this study, we have presented a symptoms-disease dataset in Bangla, collected by conducting surveys in the medical field as well as exploring online sources, to overcome the scarcity of resources available in Bangla for disease prediction. We have also provided detailed information about the diseases, symptoms, and their relations in our dataset. There are 85 unique diseases and 172 unique symptoms in our dataset along with 758 unique relationships between them. Initially, some data cleaning and preprocessing steps were carried out to predict diseases from Bangla input symptoms. Then we evaluated eight machine learning models on our dataset. Random forest, logistic regression, and perceptron were the top three models in terms of performance. Finally, we have designed soft and hard voting ensemble algorithms combining these three models and found that they provide better performance. In order to improve the dataset's inclusiveness and thoroughness, the future plan is to include all the diseases available in the medical field and increase the whole dataset to explore deep learning models as well.

\bibliographystyle{./IEEEtran}


\end{document}